%% file: main.tex
\pdfoutput=1
\documentclass{article}
\usepackage{spconf,amsmath,graphicx}
\IfFileExists{silence.sty}{\usepackage{silence}}{\providecommand\WarningFilter[2]{}}
\usepackage{microtype}  %
\usepackage{url}
\usepackage{booktabs}
\usepackage{siunitx}
\usepackage[table]{xcolor}
\usepackage{tikz}        %
\usetikzlibrary{arrows.meta}
\usepackage{pgfplots}
\pgfplotsset{compat=1.17}
\usepackage{placeins}   %
\usepackage{enumitem}
\setlist{nosep,leftmargin=13pt}
\usepackage[capitalise]{cleveref}

\usepackage{stfloats}
\usepackage{balance}   %

\input{numbers_stats.tex}

\input{numbers_timing.tex}
\input{numbers_fitinf.tex}


\title{Exemplar: Few-Shot Biomedical Segmentation That Fits in Seconds}

\name{Michal Pr\r{u}\v{s}ek$^{1,2}$ \quad Adam Novoz\'{a}msk\'{y}$^{1}$ \quad Filip \v{S}roubek$^{1}$\thanks{Code, configurations, and the score records behind every table and figure: \protect\url{https://github.com/michalprusek/Exemplar}}}
\address{
  $^{1}$The Czech Academy of Sciences,  Institute of Information Theory and Automation, Czechia\\
  $^{2}$Czech Technical University in Prague, Faculty of Nuclear Sciences and Physical Engineering, Czechia}

\begin{document}
\raggedbottom  %
\ninept
\maketitle

\begin{abstract}
Segmenting a new microscopy dataset usually means training a deep network on its annotations, and training again for every added image. Exemplar solves a segmentation readout in closed form and updates it exactly with each new mask in seconds, as accurately as networks trained for minutes. The updated readout is identical to a fit from all masks at once. Its linear readout combines frozen DINOv3 features, read at a magnification chosen from the masks, with classical filter responses at native resolution. Its regularisation and its pixel weighting are chosen exactly by leaving out one annotated image at a time, so none is set aside for validation. On ten biomedical datasets from cells to vessels and filaments, Exemplar is more accurate than every few-shot method that only applies a pretrained model, and on average than nnU-Net trained on the same masks. A fine-tuned Cellpose-SAM, the strongest baseline, is as accurate on average, but each of its fits takes a median of \cpFitMedMin{} minutes against \exFitMed{} seconds, and each new mask means fine-tuning it again.
\end{abstract}


\section{Introduction}
\label{sec:intro}

\begin{figure*}[!t]
\centering
\input{fig_architecture.tex}
\caption{Overview of Exemplar. Frozen DINOv3 features in a coarse and a magnified fine view and classical priors are combined by a linear readout solved in closed form. Shown: a Decay test image and Exemplar's prediction.}
\label{fig:architecture}
\end{figure*}

A biologist segmenting structures in a new microscopy dataset faces a practical trade-off. Cellular specialists such as Cellpose~\cite{cellpose,cellposesam}, StarDist~\cite{stardist}, and micro-SAM~\cite{microsam} produce excellent instance masks for cell-like objects. Their object models are built for cells, and retargeting them to vessels, membranes, or filaments means further annotation and training. Promptable foundation models such as the Segment Anything family~\cite{sam,sam3} are more general. They are steered by prompts at inference time, which in SAM~3 include box prompts in the image itself, rather than adapted to a dataset-specific target from a small labelled set. Interactive pixel classifiers such as ilastik~\cite{ilastik} fit a random forest to a few painted labels on classical image features, which describe local image properties but not the representations learned by pretrained networks. Interactive microscopy tools now use pretrained embeddings as features for a shallow classifier~\cite{convpaint,featureforest}. Docherty et al.\ combine both types of features in such a classifier for micrographs, trained on sparse labels, after upsampling the embeddings with a network trained in advance~\cite{docherty2025unet}. Where the two types of features have been compared directly for microscopy pixel classification, the learned features were reported to be the stronger~\cite{teuber2026vfm}. Frozen DINOv3 features have also been read at several depths~\cite{segdino,dinomvr} and input resolutions~\cite{dinomvr} by small trained probes for medical segmentation, where a non-linear probe outperformed a linear one~\cite{dinomvr}.

Few-shot segmentation offers a middle path between retraining a network and prompting every image, and it matches how microscopy datasets are often annotated. The user provides a small support set, $K$ images with complete foreground masks, and the model segments the remaining, visually similar images from the same dataset. The task is personalisation to a dataset-specific target rather than generalisation to an unseen class. Among existing methods, INSID3 relies on feature correspondence~\cite{insid3}, while SegGPT treats segmentation as painting the target regions~\cite{seggpt}. UniverSeg and Tyche are networks trained across many medical segmentation tasks~\cite{universeg,tyche}. All of these apply a pretrained model to the support set in a single forward pass and fit nothing to the new dataset.

We target the use in which a biologist annotates a few images of a new dataset and the model is refitted each time another image is annotated. There, the annotator corrects the model's prediction on the next image instead of drawing its mask from scratch, and each correction waits for a refit. The time of a fit therefore matters beside its accuracy. We introduce Exemplar, which reads frozen learned features and classical image priors at every pixel, learns from a few complete masks, and interpolates the embeddings to native resolution without an upsampling network. Exemplar solves a segmentation readout in closed form and updates it exactly with each new mask in seconds, as accurately as networks trained for minutes. A fit therefore needs no gradient descent, learning rate or stopping rule. It stays linear because only then are its left-out errors and its updates exact. Linear readouts of frozen DINOv3 features are a standard probe~\cite{dinov3}, and leaving out or adding a sample is exact for ridge regression~\cite{allen1974press,ranpac}. Our contribution is what makes such a readout accurate on microscopy from a few masks. Its sums over every pixel are computed exactly from the patch grids, and its fine view is magnified as far as the annotated structures require. A blend of two pixel weightings replaces the search for a threshold. Finally, whole images are left out, and each fold recomputes its standardisation and class weights from the stored sums. We compare Exemplar with forward-pass few-shot methods, with nnU-Net~\cite{nnunet} trained from scratch on the same masks, a baseline that few-shot comparisons rarely include, and with cell specialists fine-tuned on them. A fit takes Exemplar a median of \exFitMed{} seconds and adding a mask \exRefitLo{} to \exRefitHi{} seconds, whereas a trained network takes minutes to hours.

\section{Method}
\label{sec:method}

\noindent\textbf{Coarse and fine views.} Exemplar combines learned features from a frozen DINOv3 backbone~\cite{dinov3} with classical filter responses in a linear readout, the only part that is fitted (\cref{fig:architecture}). DINOv3 ViT-L/16 returns one feature vector per $16{\times}16$-pixel patch of its input, so one view of a large image cannot provide both context and detail. We therefore read every image twice. For the coarse view, the whole image is resized to the $672{\times}672$-pixel encoder input. For the fine view, the image is magnified by an integer factor $z$ and encoded in overlapping 672-pixel windows, so that each patch covers $16/z$ pixels of the original image, within a cap given below. Each view is represented by the concatenated feature vectors of the middle (twelfth) and the last layer of the backbone, 1024 channels each and 2048 in all. Using both layers improves the object and the thin-structure datasets over the last layer alone (\cref{tab:ablation}).

\noindent\textbf{Magnification from the support masks.} A patch cannot describe a structure narrower than itself, and the annotated structures range from vessels a few pixels wide to whole spheroids. A vessel three pixels wide, for example, fills a small part of a 16-pixel patch, but most of a patch when the image is magnified four times. The magnification $z$ is therefore set at every fit from the support masks alone and used for every image. The rule asks how well each support mask can be approximated on the magnified patch grid when each patch takes a single label. Plain Dice would judge such an approximation by the thick parts of a structure and overlook its thin branches. Each pixel is therefore weighted by $1/t$, where $t$ is the local thickness of the annotated structure at its nearest centreline point, so that a structure counts by its length rather than its area. The magnification is the smallest factor at which the best such approximation reaches a weighted Dice of 0.8 against the masks, in the median over the support, a value chosen in experiments during development (\cref{sec:exp}). The cost of the fine view grows with $z^2$, so the factor is at most 4 and at most 8192 divided by the image's longest side. On images longer than 2560 pixels, the fine grid is resampled to $160z$ patches per side.

\noindent\textbf{Classical image priors.} Patch features describe the image only at the spacing of their grid, even when it is magnified, so boundaries and fine texture are blurred. A fixed filter bank supplies this detail at native resolution. It reads the grayscale image, normalised per image. It produces 35 response maps in six families: Frangi vesselness for tubes~\cite{frangi}, Laplacian-of-Gaussian responses for blobs, Sauvola adaptive thresholds for local contrast, structure-tensor eigenvalues for edges and anisotropy, gradient magnitude, and normalised intensity. The bank comes from our earlier work~\cite{hyperbank}.

\noindent\textbf{Linear readout in closed form.} For each image, the inputs thus form a stack of 4132 maps at native resolution. They are the 2048 channels of each view, interpolated bilinearly from its grid, the 35 prior responses, and a scalar constant that plays the role of a bias. Let the vector $f_p$ denote the values of these maps at pixel $p$. Each map except the constant is standardised over the support pixels, so that one penalty treats all of them alike. The readout scores the pixel as $s_p = w^{\top}f_p$, that is, a weighted sum of the maps. The weights $w$ are estimated against the support masks $y_p \in \{0,1\}$ by weighted least squares, regularised with a ridge penalty:
\begin{equation}
w = \arg\min_{w}\, \frac{1}{n}\sum_{p} q_p\,(y_p - w^{\top}f_p)^2 + \lambda\, w^{\top} I_0\, w.
\label{eq:ridge}
\end{equation}
The solution is $w = (A/n + \lambda I_0)^{-1}\, b/n$. Here $q_p$ is the weight of pixel $p$, $A = \sum_p q_p f_p f_p^{\top}$ the weighted Gram matrix, $b = \sum_p q_p y_p f_p$ and $n = \sum_p q_p$. The matrix $I_0$ is the identity with a zero for the constant, which is therefore not penalised. We use the squared error because the weights then solve a single linear system and the left-out error below is exact.

\noindent\textbf{Computing on the patch grids.} For one view, let the matrix $F$ hold its patch features, one row per patch, and let $U$ be the bilinear interpolation from the patches to the pixels. Row $p$ of $UF$ is then the view's part of $f_p$ before standardisation. Exemplar computes $A$ and $b$ without forming $UF$ for whole images, because its 2048 maps at native resolution would take much memory. Two properties make this possible and limit the interpolation to the foreground pixels. First, each weighting below gives $q_p$ one value on the foreground and one on the background. Each support image $k$ therefore keeps only its raw sums, unweighted and unstandardised, over all its pixels and over its foreground (their difference gives the background), and any weighting and standardisation of its contributions $A_k$, $b_k$ follow from them algebraically. Second, bilinear interpolation is linear. With $Q$ the diagonal matrix of the weights $q_p$, the view's block of $A$ is therefore
\begin{equation}
A_F = (UF)^{\top} Q\, UF = F^{\top} (U^{\top} Q\, U)\, F.
\label{eq:gridgram}
\end{equation}
By the first property, $U^{\top} Q\, U$ combines $U^{\top} U$, which depends only on the image and grid sizes and is computed on the grid, with the same product over the foreground pixels alone. Only at these pixels are the features interpolated. The other blocks of $A$, and $b$, are formed in the same way.

\noindent\textbf{Two weightings and their blend.} We solve \cref{eq:ridge} twice, with two weightings. The plain readout $w_{\mathrm{plain}}$ sets $q_p = 1$ for every pixel, so its score estimates the probability of foreground. Thin structures, however, cover only a few percent of an image, so their pixels often stay below one half. The balanced readout $w_{\mathrm{bal}}$ therefore gives foreground and background equal total weight, with $q_p = 1/(2\rho)$ on the foreground and $1/(2(1-\rho))$ on the background, $\rho$ being the foreground fraction of the support pixels. Thresholded at one half, it keeps roughly every pixel whose foreground probability exceeds $\rho$, so thin structures survive but compact objects grow. Exemplar then blends the two readouts into $w_a = a\,w_{\mathrm{plain}} + (1-a)\,w_{\mathrm{bal}}$ and thresholds its score at one half. A blend changes both the relative weights of the features and how much of the image passes the threshold, so this one number replaces both the choice of weighting and a search for the threshold.

\noindent\textbf{Choosing the blend.} The blend $a$ is chosen from the support masks alone, by leaving out each support image in turn, or each quadrant of a single one, and predicting it from the rest (\cref{eq:loo}). Such a left-out map is linear in the weights, so that of any blend is the same blend of the two, and choosing $a$ needs no further fit or feature interpolation. We try $a$ from 0 to 1 in steps of 0.2 and then up to four values around the best. Each left-out map is scored against its mask by the dataset's own metric, the only input given per dataset. The plain readout ($a = 1$) is kept unless a blend is better on the left-out images by more than one standard error of the per-image difference. A nearly flat left-out curve thus cannot lead the choice into noise.

\noindent\textbf{Ridge weight from held-out support images.} The ridge weight $\lambda$ is chosen by the same leaving out. At the same magnification, the fit without image $k$ is a fit from scratch on the other images, whose raw sums give the mean and standard deviation of every map, the foreground fraction and so the class weights. All sums below, those of image $k$ included, are taken with this standardisation and these weights. Its squared error $E_k$ on image $k$ follows from that image's own sums:
\begin{equation}
\begin{gathered}
w_{-k} = \big((A - A_k)/(n - n_k) + \lambda I_0\big)^{-1}\,(b - b_k)/(n - n_k),\\
E_k = \big(c_k - 2\,w_{-k}^{\top} b_k + w_{-k}^{\top} A_k\, w_{-k}\big)/n_k,
\end{gathered}
\label{eq:loo}
\end{equation}
where $n_k$ is the total weight of image $k$ and $c_k = \sum_{p \in k} q_p y_p$ its weighted foreground. Choosing $\lambda$ therefore needs no pass over the pixels. For each weighting, $\lambda$ minimises the mean of $E_k$ over the images. Unlike Dice or IoU, this error is smooth in $\lambda$, so we find its minimum by Newton's method on $\log\lambda$.

\noindent\textbf{Prediction.} A new image is segmented by thresholding the score map of the chosen blend at one half. With $w_F$ the weights of a view, $UF\,w_F = U(F\,w_F)$, so each view is weighted on its grid and only the resulting map is interpolated.

\noindent\textbf{Adding a mask.} A newly masked image contributes one more pair $A_k$, $b_k$. The earlier images are not encoded again, because their grids and sums are kept; only a change of $z$ forces their fine views to be recomputed. The standardisation and every choice are then made from the updated sums as a fit from scratch would make them, so the result is identical to a fit from all masks at once.

\begin{figure}[tb]
\centering
\input{fig_kscale.tex}
\caption{Mean score, as in \cref{tab:fewshot}, against the number of support masks $K$. Bands: 95\% confidence intervals over seeds.}
\label{fig:kscale}
\end{figure}

\begin{figure}[tb]
\centering
\setlength{\fboxsep}{0pt}\setlength{\fboxrule}{0.4pt}%
\newcommand{\qq}[2]{\begin{minipage}[t]{0.235\linewidth}\centering\fbox{\includegraphics[width=\dimexpr\linewidth-0.8pt\relax]{#1}}\\[0.5pt]{#2}\end{minipage}}
\qq{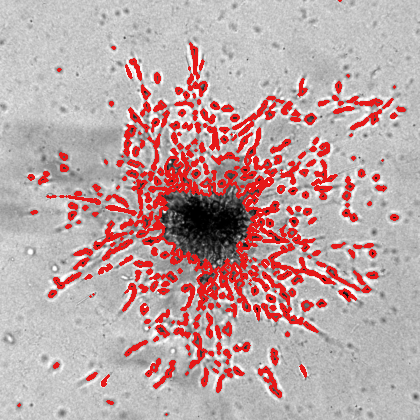}{Decay}\hfill\qq{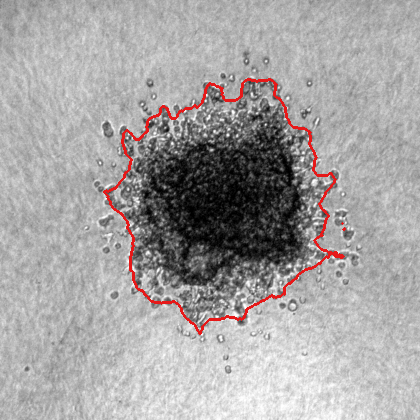}{SpheroidJ}\hfill\qq{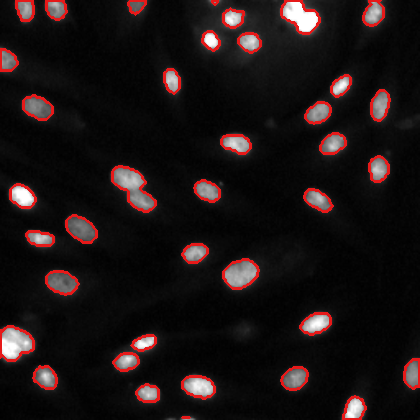}{DSB2018}\hfill\qq{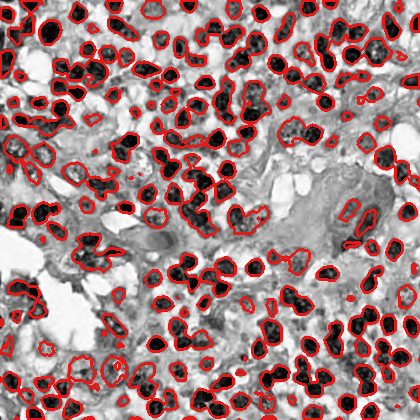}{MoNuSeg}\\[3pt]
\qq{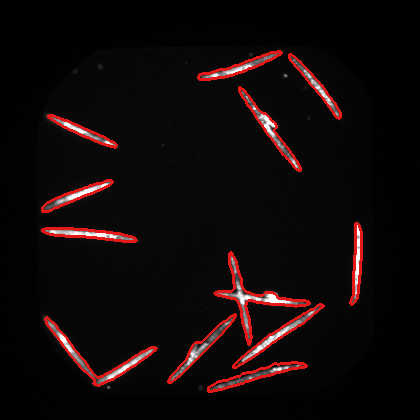}{BBBC010}\hfill\qq{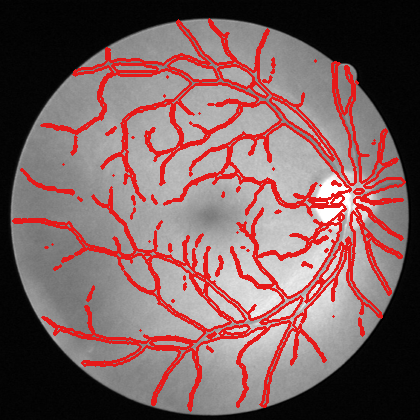}{DRIVE}\hfill\qq{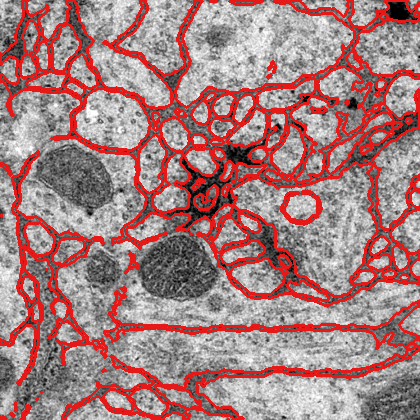}{ISBI2012-EM}\hfill\qq{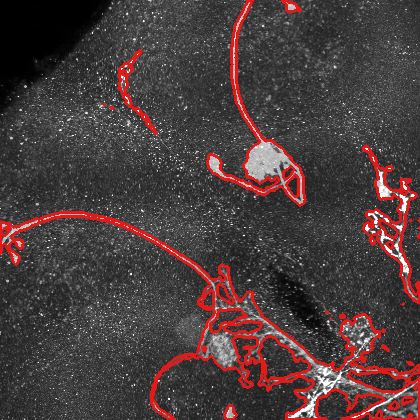}{FISBe}
\caption{Exemplar's predictions (red contours) with eight masks, on one test image of each of eight datasets.}
\label{fig:qual}
\end{figure}

\section{Experiments}
\label{sec:exp}

\noindent\textbf{Datasets and protocol.} We evaluate on ten biomedical imaging datasets (\cref{fig:qual}). The object datasets hold spheroid and nucleus blobs (SpheroidJ~\cite{spheroidj}, DSB2018~\cite{dsb2018}, MoNuSeg~\cite{monuseg}), decaying spheroids (Decay~\cite{hyperbank}, released with that work), overlapping \emph{C.\,elegans} bodies (BBBC010~\cite{bbbc010}) and densely packed bacteria in phase contrast (Bacteria~\cite{omnipose}). The thin-structure datasets hold retinal vessels (DRIVE~\cite{drive}, HRF~\cite{hrf}), electron-microscopy neuronal membranes (ISBI2012-EM~\cite{isbi2012em}) and fluorescent filaments (FISBe~\cite{fisbe}). We score the object datasets by foreground intersection-over-union (IoU) and the thin-structure datasets by centreline Dice~\cite{cldice}, which is less sensitive than overlap to errors in the width of structures a few pixels wide. Each of ten seeds draws $K$ support masks at random from a fixed pool of 16 to 20 annotated images. The test images, 14 to 148 per dataset, are disjoint from the pool and the same for every $K$ and every method. No separate validation datasets were held out. We set the constants of the method while developing it on these datasets, test images included: the middle (twelfth) backbone layer, the weighted Dice of 0.8 and the cap $z \le 4$ of the magnification rule, and the blend search steps. Each constant has one value for all datasets and all numbers of masks, and none was tuned per dataset. The score of a dataset is its mean over test images and seeds, and a mean over datasets weights each dataset equally. We compare methods with paired Wilcoxon signed-rank tests on per-image scores averaged over the seeds. Holm correction is applied within each group of comparisons: the \fpComparisons{} with the four forward-pass methods, and each of nnU-Net, fine-tuned Cellpose-SAM and Convpaint against Exemplar at eight masks.

\noindent\textbf{Baselines.} We compare with three groups of methods. The first group comprises \emph{forward-pass} few-shot methods, which apply a pretrained model to the support set without fitting: UniverSeg~\cite{universeg}, Tyche~\cite{tyche}, INSID3~\cite{insid3} with the CRF refinement of its published results, and SegGPT~\cite{seggpt}. Tyche is scored, as in its paper, by its best candidate against the test mask, here the best of 16. The second group is fitted without gradient descent. The pixel classifier ilastik~\cite{ilastik} trains its default random forest on all of its filter features, using pixels sampled from the support masks. Convpaint~\cite{convpaint} trains boosted trees on the features of two pretrained networks, in the variant that scored best on test images. The third group trains networks on the support set. nnU-Net~\cite{nnunet,nnunetrevisited} is trained from scratch in its 2D residual-encoder configuration for 100 epochs, a length chosen on test images. We chose it because, on one support draw, its default of 1000 epochs raised the test score on one of four datasets and lowered it on the other three. The specialists StarDist~\cite{stardist}, Cellpose-SAM~\cite{cellposesam} and micro-SAM~\cite{microsam} use no support masks as released, so we fine-tune each on the same masks, with a training length selected on test images. Without fine-tuning, they score far lower, especially on the thin structures, because their object models are built for cells. Each specialist is scored on its own foreground map, because the union of its objects misses structures that its object model cannot represent. Every choice made for a baseline on test images favours it. All groups are compared at matched annotation budget, not computation.

\section{Results}
\label{sec:results}

\input{tab_fewshot.tex}

\noindent\textbf{Fitting a readout beats conditioning a pretrained model.} Exemplar is ahead of every forward-pass method on every dataset in \cref{tab:fewshot}, significantly in \fpSig{} of the \fpComparisons{} comparisons, and at every number of masks in \cref{fig:kscale}. As released, UniverSeg and Tyche see each image at 128 pixels, too coarse for thin structures on large images. Convpaint, which also fits a classifier to pretrained features of the support pixels, is ahead of the forward-pass methods on average but significantly behind Exemplar on every dataset.

\noindent\textbf{On average, Exemplar is more accurate than nnU-Net trained on the same masks, at every number of masks.} With eight masks, Exemplar's mean is higher than that of nnU-Net on both groups of datasets (\cref{tab:fewshot}). Exemplar is ahead on \nnOursAheadWord{} of the ten datasets, significantly on \oursSigDs{}. nnU-Net is ahead on the other \nnAheadWord{}, significantly on \nnSigDs{}. Fine-tuned Cellpose-SAM and Exemplar have the highest means of all methods, and the two are level. Cellpose-SAM is significantly ahead on \cpSigWord{} datasets and Exemplar on \cpOursSigWord{}. On the datasets scored by centreline Dice, Exemplar has the highest mean of all methods. In \cref{fig:kscale}, the gap to nnU-Net is largest with few masks and is still a small lead at the largest number of masks. Fine-tuned Cellpose-SAM stays close to Exemplar at every number of masks and pulls slightly ahead at the largest.

\noindent\textbf{Exemplar fits and updates in seconds, where trained networks need minutes.} Exemplar fits in \exFitLo{} to \exFitHi{} seconds, the longest on the largest magnified images, and segments an image in \exInfLo{} to \exInfHi{} seconds, while every trained network takes minutes to fit (medians in \cref{tab:fewshot}). Convpaint and ilastik, also fitted without gradient descent, take several times longer than Exemplar and must be refitted from scratch after each new mask. Segmenting an image takes Exemplar longer than the fine-tuned specialists, because it encodes the image in both views and computes the priors at full resolution. Adding the eighth mask to a fit of seven is exact and takes \exRefitLo{} to \exRefitHi{} seconds. A trained network has no exact update. Trained further for only an eighth of its training length after one more mask, Cellpose-SAM takes \cpEighthLo{} to \cpEighthHi{} seconds.

\input{tab_ablation.tex}

\noindent\textbf{Every input except the coarse view contributes clearly, and no fixed readout serves both groups of datasets.} \Cref{tab:ablation} changes one part of Exemplar at a time and solves the readout again on the same masks. Without the priors, the mean falls on the datasets scored by overlap, while those scored by centreline Dice barely change. Without the fine view, the thin structures lose most, and reading the fine view without magnification costs them nearly as much. The coarse view adds little: without it, the datasets scored by overlap lose slightly and those scored by centreline Dice gain slightly. We keep it because it is the readout's only whole-image context: at $z = 4$, a fine window covers 168 pixels of the original image. Of the two ends of the blend, plain at one half loses most on the datasets scored by centreline Dice, and balanced at one half on those scored by overlap. The blend chosen on held-out images beats both ends in both groups and adds about as much to the mean as the priors. A random forest on the same inputs is less accurate than the linear readout on average, both on the views and on the priors. These results support the magnified fine view read at two depths beside the priors, choosing the blend on held-out support images, and keeping the readout linear.

\section{Discussion and Conclusion}
\label{sec:conc}

Exemplar solves a segmentation readout in closed form and updates it exactly with each new mask in seconds, as accurately as networks trained for minutes. With eight masks it is more accurate than nnU-Net trained from scratch on them, though significantly behind it on \nnSigDs{}, and its mean stays above nnU-Net's from one to sixteen masks. Fine-tuned Cellpose-SAM stays close to Exemplar at every number of masks and pulls slightly ahead at the largest. Exemplar is limited to a binary foreground in 2D images; a readout of distance transforms could split touching objects.

\FloatBarrier   %
\section{Compliance with Ethical Standards}
\label{sec:ethics}
This research was conducted retrospectively using publicly available, open-access biomedical imaging datasets. No new human or animal data were collected, and ethical approval was not required as confirmed by the licenses attached to the open-access data.

\section{Acknowledgments}
\label{sec:ack}
This document is the result of the grant GA25-15933S funded by the Czech Science Foundation. The code was written and the text edited with the assistance of Claude (Anthropic). The authors have no relevant financial or non-financial interests to disclose.

\sloppy\hbadness=2500
\let\oldbibliography\thebibliography
\renewcommand{\thebibliography}[1]{\oldbibliography{#1}%
  \setlength{\itemsep}{0pt}\setlength{\parsep}{0pt}\setlength{\parskip}{0pt}}
\balance
\bibliographystyle{IEEEbib}
\bibliography{refs}

\end{document}

%% file: numbers_stats.tex
\newcommand{\fpComparisons}{40}

\newcommand{\fpSig}{39}

\newcommand{\nnAheadWord}{three}
\newcommand{\oursSigDs}{SpheroidJ, BBBC010, DRIVE and HRF}

\newcommand{\nnSigDs}{Bacteria and ISBI2012-EM}
\newcommand{\nnOursAheadWord}{seven}

\newcommand{\cpOursSigWord}{four}

\newcommand{\cpSigWord}{five}

%% file: numbers_fitinf.tex
\newcommand{\tinfSegGPT}{2.2\,s}

\newcommand{\tinfUniverSeg}{43\,ms}

\newcommand{\tinfINSID}{1.9\,s}

\newcommand{\tinfTyche}{290\,ms}
\newcommand{\tfitExemplar}{8.5\,s}
\newcommand{\tinfExemplar}{620\,ms}

\newcommand{\tfitConvpaint}{43\,s}
\newcommand{\tinfConvpaint}{4.2\,s}
\newcommand{\tfitnnUNet}{53\,min}
\newcommand{\tinfnnUNet}{490\,ms}
\newcommand{\tfitCellposeSAM}{25\,min}
\newcommand{\tinfCellposeSAM}{490\,ms}
\newcommand{\tfitStarDist}{8\,min}
\newcommand{\tinfStarDist}{150\,ms}
\newcommand{\tfitmicroSAM}{6\,min}
\newcommand{\tinfmicroSAM}{270\,ms}
\newcommand{\tfitilastik}{40\,s}
\newcommand{\tinfilastik}{1.3\,s}
\newcommand{\exFitLo}{2.6}
\newcommand{\exFitHi}{100}
\newcommand{\exRefitLo}{0.91}
\newcommand{\exRefitHi}{18}

\newcommand{\exFitMed}{8.5}
\newcommand{\exInfLo}{0.19}
\newcommand{\exInfHi}{10}

\newcommand{\cpFitMedMin}{25}
\newcommand{\cpEighthLo}{160}
\newcommand{\cpEighthHi}{280}

%% file: fig_architecture.tex
\begin{tikzpicture}[x=1pt, y=1pt, >={Stealth[length=4.2pt, width=3.6pt]},
    every node/.style={font=\fontsize{9}{10.2}\selectfont, inner sep=1.5pt, align=center},
    box/.style={draw, rounded corners=3pt, line width=0.5pt, minimum height=23pt},
    frozen/.style={box, fill=exGreenFill, draw=exGreenLine},
    prior/.style={box, fill=exPurpleFill, draw=exPurpleLine},
    head/.style={box, fill=exRedFill, draw=exRedLine},
    fit/.style={box, fill=exYellowFill, draw=exYellowLine},
    flow/.style={->, line width=0.7pt},
    signal/.style={->, line width=0.7pt, dashed, draw=exSignal},
    lab/.style={inner sep=1pt, text=exSignal}]
  \definecolor{exGreenFill}{HTML}{D5E8D4} \definecolor{exGreenLine}{HTML}{82B366}
  \definecolor{exRedFill}{HTML}{F8CECC}   \definecolor{exRedLine}{HTML}{B85450}
  \definecolor{exPurpleFill}{HTML}{E1D5E7}\definecolor{exPurpleLine}{HTML}{9673A6}
  \definecolor{exYellowFill}{HTML}{FFF2CC}\definecolor{exYellowLine}{HTML}{D6B656}
  \definecolor{exSignal}{HTML}{B40000}

  \path[use as bounding box] (0,-39) rectangle (506,29);

  \node[inner sep=0pt, draw=black!40, line width=0.4pt] (inp) at (28,0)
    {\includegraphics[width=56pt]{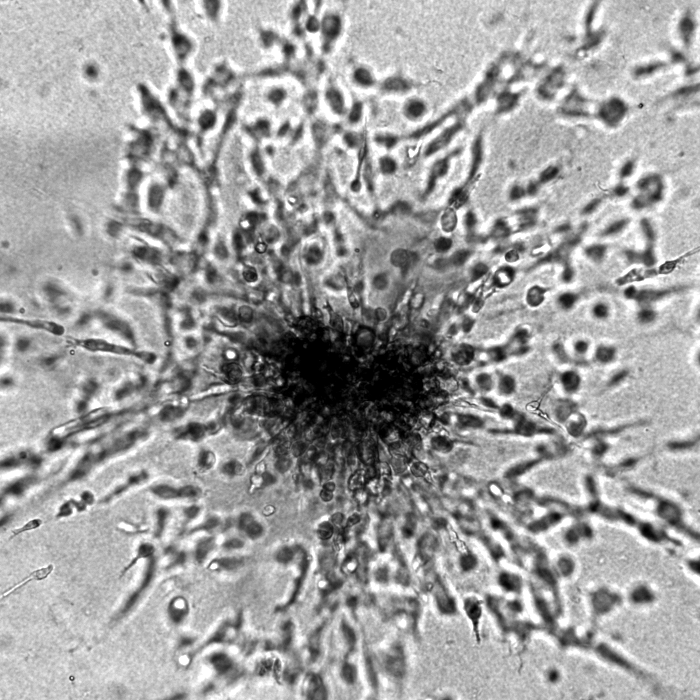}};
  \node[anchor=north, inner sep=1pt] at (28,-29) {Image};
  \node[inner sep=0pt, draw=black!40, line width=0.4pt] (out) at (478,0)
    {\includegraphics[width=56pt]{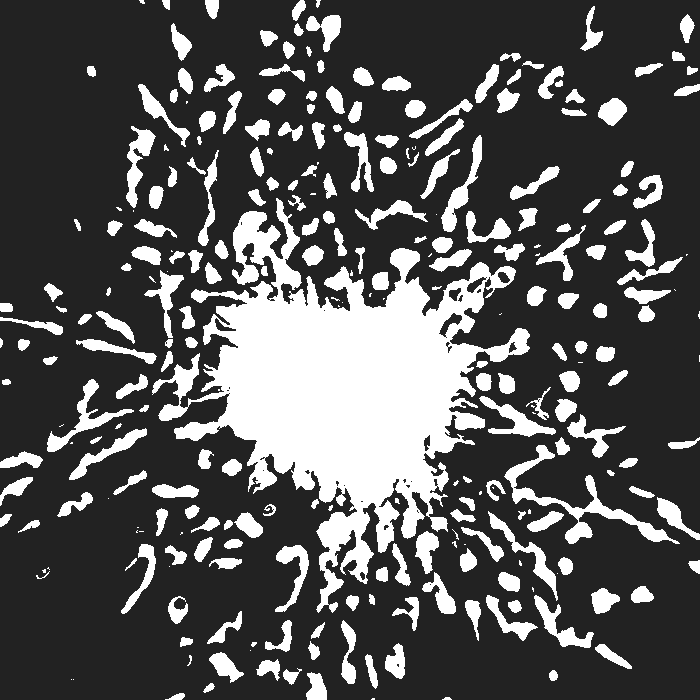}};
  \node[anchor=north, inner sep=1pt] at (478,-29) {$s > 0.5$};

  \node[frozen, minimum width=156pt] (dino) at (148,16.5)
    {\textbf{Frozen DINOv3 ViT-L/16}\\coarse + fine view, magnified $z{\times}$};
  \node[prior, minimum width=156pt] (bank) at (148,-16.5)
    {\textbf{Classical priors, 35 maps}\\Frangi, LoG, Sauvola, structure tensor, \dots};

  \node[head, minimum width=186pt] (read) at (339,16.5)
    {\textbf{Linear readout, 4132 weights}\\$s = w^{\top}f$ at every pixel, $f$: $2048{+}2048{+}35{+}1$};
  \node[fit, minimum width=186pt] (fit) at (339,-16.5)
    {\textbf{Closed form on $K$ support masks}\\leave one image out: $\lambda$, blend $a$};

  \draw[flow] (inp.east) -- (63,0) |- (dino.west);
  \draw[flow] (63,0) |- (bank.west);
  \draw[flow] (dino.east) -- (read.west);
  \draw[line width=0.7pt] (bank.east) -- (236,-16.5) -- (236,16.5);
  \fill (236,16.5) circle (1.3pt);
  \draw[flow] (read.east) -- (out.west |- read.east);

  \draw[signal] (300,-5) -- (300,5);
  \node[lab, anchor=west] at (303,0) {$w$};
\end{tikzpicture}%

%% file: fig_kscale.tex
\begin{tikzpicture}[font=\fontsize{9}{10.2}\selectfont]
\definecolor{ksA}{HTML}{111111}
\definecolor{ksB}{HTML}{C7522B}
\definecolor{ksC}{HTML}{9467BD}
\definecolor{ksD}{HTML}{1F77B4}
\definecolor{ksE}{HTML}{2CA02C}
\definecolor{ksF}{HTML}{8C564B}
\definecolor{ksG}{HTML}{17BECF}
\definecolor{ksH}{HTML}{BCBD22}
\definecolor{ksI}{HTML}{E377C2}
\begin{axis}[scale only axis, width=\dimexpr\linewidth-108pt\relax, height=66pt,
  xmode=log, log basis x=2, xmin=0.871, xmax=18.379, xtick={1,4,8,16}, xticklabels={1,4,8,16},
  xminorticks=false, ymin=0.2, ymax=0.871, ytick={0.2,0.4,0.6,0.8},
  yticklabel style={/pgf/number format/fixed, /pgf/number format/precision=1},
  xlabel={Support masks $K$}, ylabel={Mean score},
  xlabel style={yshift=2pt}, ylabel style={yshift=-3pt},
  grid=major, major grid style={dotted, black!35}, tick align=outside, tick pos=left,
  legend style={at={(1.03,0.5)}, anchor=west, draw=none, fill=none, inner sep=0pt,
    row sep=-1.5pt, legend cell align=left}, legend image post style={xscale=0.7}]
\addplot[draw=none, fill=ksB!15, forget plot] coordinates {(1,0.6533) (4,0.7576) (8,0.7993) (16,0.8167) (16,0.8234) (8,0.8089) (4,0.7781) (1,0.6933)} -- cycle;
\addplot[draw=none, fill=ksC!15, forget plot] coordinates {(1,0.4556) (4,0.4968) (8,0.5021) (16,0.5088) (16,0.5137) (8,0.5131) (4,0.5146) (1,0.4927)} -- cycle;
\addplot[draw=none, fill=ksD!15, forget plot] coordinates {(1,0.3958) (4,0.5192) (8,0.5464) (16,0.5652) (16,0.5681) (8,0.5535) (4,0.5304) (1,0.4113)} -- cycle;
\addplot[draw=none, fill=ksE!15, forget plot] coordinates {(1,0.2182) (4,0.3862) (8,0.4504) (16,0.4912) (16,0.4938) (8,0.4630) (4,0.4017) (1,0.2312)} -- cycle;
\addplot[draw=none, fill=ksF!15, forget plot] coordinates {(1,0.4022) (4,0.4322) (8,0.4297) (16,0.4304) (16,0.4351) (8,0.4342) (4,0.4392) (1,0.4298)} -- cycle;
\addplot[draw=none, fill=ksG!15, forget plot] coordinates {(1,0.7606) (4,0.7998) (8,0.8145) (16,0.8344) (16,0.8371) (8,0.8310) (4,0.8128) (1,0.7873)} -- cycle;
\addplot[draw=none, fill=ksH!15, forget plot] coordinates {(1,0.5840) (4,0.6900) (8,0.7324) (16,0.7465) (16,0.7507) (8,0.7394) (4,0.7143) (1,0.6409)} -- cycle;
\addplot[draw=none, fill=ksI!15, forget plot] coordinates {(1,0.6952) (4,0.7180) (8,0.7304) (16,0.7511) (16,0.7617) (8,0.7431) (4,0.7357) (1,0.7125)} -- cycle;
\addplot[draw=none, fill=ksA!15, forget plot] coordinates {(1,0.7633) (4,0.8049) (8,0.8191) (16,0.8272) (16,0.8287) (8,0.8249) (4,0.8121) (1,0.7838)} -- cycle;
\addlegendimage{ksA, line width=1.55pt, dashed, mark=none}
\addlegendentry{Exemplar}
\addplot[ksB, line width=1.29pt, mark=none] coordinates {(1,0.6733) (4,0.7679) (8,0.8041) (16,0.8201)};
\addlegendentry{nnU-Net}
\addplot[ksC, line width=0.90pt, mark=none] coordinates {(1,0.4742) (4,0.5057) (8,0.5076) (16,0.5113)};
\addlegendentry{SegGPT}
\addplot[ksD, line width=0.90pt, mark=none] coordinates {(1,0.4036) (4,0.5248) (8,0.5500) (16,0.5667)};
\addlegendentry{Tyche}
\addplot[ksE, line width=0.90pt, mark=none] coordinates {(1,0.2247) (4,0.3939) (8,0.4567) (16,0.4925)};
\addlegendentry{UniverSeg}
\addplot[ksF, line width=0.90pt, mark=none] coordinates {(1,0.4160) (4,0.4357) (8,0.4320) (16,0.4328)};
\addlegendentry{INSID3}
\addplot[ksG, line width=0.90pt, mark=none] coordinates {(1,0.7739) (4,0.8063) (8,0.8227) (16,0.8357)};
\addlegendentry{Cellpose-SAM}
\addplot[ksH, line width=0.90pt, mark=none] coordinates {(1,0.6125) (4,0.7021) (8,0.7359) (16,0.7486)};
\addlegendentry{StarDist}
\addplot[ksI, line width=0.90pt, mark=none] coordinates {(1,0.7039) (4,0.7269) (8,0.7367) (16,0.7564)};
\addlegendentry{micro-SAM}
\addplot[ksA, line width=1.55pt, dashed, mark=none, forget plot] coordinates {(1,0.7735) (4,0.8085) (8,0.8220) (16,0.8279)};
\end{axis}
\end{tikzpicture}%

%% file: tab_fewshot.tex
\begin{table*}[t]
\centering
\caption{Scores with $K{=}8$ support masks, mean $\pm$ standard deviation over 10 seeds. The upper datasets are scored by foreground IoU (overlap), the lower by centreline Dice. Bold: best in each group of methods; underline: best in the row. The Mean averages all ten datasets.  Last row: median over the datasets of the fit time\,/\,time per image on one RTX A5000 (first support draw, five test images), encoding included.}
\label{tab:fewshot}
{\scriptsize\setlength{\tabcolsep}{0.8pt}
\begin{tabular}{l cccc ccc cccc}
\toprule
 & \multicolumn{4}{c}{Forward pass, nothing fitted} & \multicolumn{3}{c}{Fitted without gradient descent} & \multicolumn{4}{c}{Network trained on the support} \\
\cmidrule(lr){2-5} \cmidrule(lr){6-8} \cmidrule(lr){9-12}
Dataset & \begin{tabular}{@{}c@{}}SegGPT\\[-1.5pt]\cite{seggpt}\end{tabular} & \begin{tabular}{@{}c@{}}UniverSeg\\[-1.5pt]\cite{universeg}\end{tabular} & \begin{tabular}{@{}c@{}}INSID3\\[-1.5pt]\cite{insid3}\end{tabular} & \begin{tabular}{@{}c@{}}Tyche\\[-1.5pt]\cite{tyche}\end{tabular} & \begin{tabular}{@{}c@{}}\textbf{Exemplar}\\[-1.5pt](ours)\end{tabular} & \begin{tabular}{@{}c@{}}ilastik\\[-1.5pt]\cite{ilastik}\end{tabular} & \begin{tabular}{@{}c@{}}Convpaint\\[-1.5pt]\cite{convpaint}\end{tabular} & \begin{tabular}{@{}c@{}}nnU-Net\\[-1.5pt]\cite{nnunet}\end{tabular} & \begin{tabular}{@{}c@{}}Cellpose-SAM\\[-1.5pt]\cite{cellposesam}\end{tabular} & \begin{tabular}{@{}c@{}}StarDist\\[-1.5pt]\cite{stardist}\end{tabular} & \begin{tabular}{@{}c@{}}micro-SAM\\[-1.5pt]\cite{microsam}\end{tabular} \\
\midrule
SpheroidJ    & {\boldmath$0.895$}{\scriptsize$\pm.014$} & $0.804${\scriptsize$\pm.041$} & $0.877${\scriptsize$\pm.015$} & $0.863${\scriptsize$\pm.012$} & \underline{{\boldmath$0.922$}{\scriptsize$\pm.029$}} & $0.675${\scriptsize$\pm.032$} & $0.798${\scriptsize$\pm.026$} & $0.822${\scriptsize$\pm.069$} & $0.864${\scriptsize$\pm.110$} & $0.660${\scriptsize$\pm.042$} & {\boldmath$0.865$}{\scriptsize$\pm.091$} \\
Decay        & $0.498${\scriptsize$\pm.003$} & $0.490${\scriptsize$\pm.006$} & $0.464${\scriptsize$\pm.013$} & {\boldmath$0.535$}{\scriptsize$\pm.005$} & {\boldmath$0.795$}{\scriptsize$\pm.005$} & $0.629${\scriptsize$\pm.044$} & $0.679${\scriptsize$\pm.018$} & $0.805${\scriptsize$\pm.022$} & \underline{{\boldmath$0.819$}{\scriptsize$\pm.007$}} & $0.681${\scriptsize$\pm.017$} & $0.466${\scriptsize$\pm.062$} \\
DSB2018      & {\boldmath$0.777$}{\scriptsize$\pm.006$} & $0.588${\scriptsize$\pm.033$} & $0.604${\scriptsize$\pm.014$} & $0.669${\scriptsize$\pm.022$} & {\boldmath$0.859$}{\scriptsize$\pm.004$} & $0.809${\scriptsize$\pm.022$} & $0.799${\scriptsize$\pm.015$} & $0.851${\scriptsize$\pm.009$} & \underline{{\boldmath$0.870$}{\scriptsize$\pm.003$}} & $0.831${\scriptsize$\pm.005$} & \underline{{\boldmath$0.870$}{\scriptsize$\pm.005$}} \\
MoNuSeg      & $0.127${\scriptsize$\pm.030$} & $0.262${\scriptsize$\pm.021$} & $0.220${\scriptsize$\pm.001$} & {\boldmath$0.425$}{\scriptsize$\pm.005$} & {\boldmath$0.673$}{\scriptsize$\pm.010$} & $0.517${\scriptsize$\pm.044$} & $0.593${\scriptsize$\pm.014$} & $0.671${\scriptsize$\pm.008$} & \underline{{\boldmath$0.705$}{\scriptsize$\pm.006$}} & $0.671${\scriptsize$\pm.013$} & $0.689${\scriptsize$\pm.006$} \\
BBBC010      & {\boldmath$0.419$}{\scriptsize$\pm.003$} & $0.213${\scriptsize$\pm.027$} & $0.305${\scriptsize$\pm.007$} & $0.340${\scriptsize$\pm.022$} & \underline{{\boldmath$0.663$}{\scriptsize$\pm.003$}} & $0.443${\scriptsize$\pm.034$} & $0.477${\scriptsize$\pm.024$} & $0.645${\scriptsize$\pm.010$} & {\boldmath$0.653$}{\scriptsize$\pm.010$} & $0.583${\scriptsize$\pm.022$} & $0.572${\scriptsize$\pm.020$} \\
Bacteria     & $0.622${\scriptsize$\pm.036$} & $0.555${\scriptsize$\pm.028$} & {\boldmath$0.632$}{\scriptsize$\pm.004$} & $0.550${\scriptsize$\pm.044$} & {\boldmath$0.916$}{\scriptsize$\pm.009$} & $0.854${\scriptsize$\pm.033$} & $0.879${\scriptsize$\pm.008$} & $0.927${\scriptsize$\pm.011$} & \underline{{\boldmath$0.944$}{\scriptsize$\pm.005$}} & $0.907${\scriptsize$\pm.017$} & $0.887${\scriptsize$\pm.023$} \\
\midrule
\textit{Overlap} & $0.556${\scriptsize$\pm.009$} & $0.485${\scriptsize$\pm.014$} & $0.517${\scriptsize$\pm.004$} & {\boldmath$0.564$}{\scriptsize$\pm.008$} & {\boldmath$0.805$}{\scriptsize$\pm.006$} & $0.655${\scriptsize$\pm.016$} & $0.704${\scriptsize$\pm.007$} & $0.787${\scriptsize$\pm.014$} & \underline{{\boldmath$0.809$}{\scriptsize$\pm.019$}} & $0.722${\scriptsize$\pm.010$} & $0.725${\scriptsize$\pm.018$} \\
\midrule
DRIVE        & $0.400${\scriptsize$\pm.005$} & $0.349${\scriptsize$\pm.019$} & $0.222${\scriptsize$\pm.007$} & {\boldmath$0.519$}{\scriptsize$\pm.005$} & \underline{{\boldmath$0.820$}{\scriptsize$\pm.002$}} & $0.696${\scriptsize$\pm.010$} & $0.598${\scriptsize$\pm.019$} & {\boldmath$0.808$}{\scriptsize$\pm.004$} & $0.791${\scriptsize$\pm.008$} & $0.790${\scriptsize$\pm.006$} & $0.772${\scriptsize$\pm.011$} \\
HRF          & $0.235${\scriptsize$\pm.006$} & $0.126${\scriptsize$\pm.010$} & $0.175${\scriptsize$\pm.006$} & {\boldmath$0.238$}{\scriptsize$\pm.014$} & \underline{{\boldmath$0.818$}{\scriptsize$\pm.001$}} & $0.630${\scriptsize$\pm.016$} & $0.540${\scriptsize$\pm.008$} & {\boldmath$0.795$}{\scriptsize$\pm.005$} & $0.788${\scriptsize$\pm.009$} & $0.685${\scriptsize$\pm.015$} & $0.509${\scriptsize$\pm.049$} \\
ISBI2012-EM  & $0.613${\scriptsize$\pm.022$} & $0.765${\scriptsize$\pm.010$} & $0.347${\scriptsize$\pm.012$} & {\boldmath$0.860$}{\scriptsize$\pm.010$} & {\boldmath$0.936$}{\scriptsize$\pm.002$} & $0.780${\scriptsize$\pm.009$} & $0.813${\scriptsize$\pm.002$} & \underline{{\boldmath$0.947$}{\scriptsize$\pm.002$}} & $0.936${\scriptsize$\pm.004$} & $0.924${\scriptsize$\pm.005$} & $0.936${\scriptsize$\pm.004$} \\
FISBe        & $0.490${\scriptsize$\pm.011$} & $0.417${\scriptsize$\pm.015$} & $0.472${\scriptsize$\pm.022$} & {\boldmath$0.500$}{\scriptsize$\pm.014$} & {\boldmath$0.818$}{\scriptsize$\pm.010$} & $0.412${\scriptsize$\pm.058$} & $0.507${\scriptsize$\pm.046$} & $0.770${\scriptsize$\pm.029$} & \underline{{\boldmath$0.858$}{\scriptsize$\pm.009$}} & $0.625${\scriptsize$\pm.021$} & $0.804${\scriptsize$\pm.017$} \\
\midrule
\textit{Centreline} & $0.435${\scriptsize$\pm.008$} & $0.414${\scriptsize$\pm.009$} & $0.304${\scriptsize$\pm.005$} & {\boldmath$0.529$}{\scriptsize$\pm.003$} & \underline{{\boldmath$0.848$}{\scriptsize$\pm.003$}} & $0.629${\scriptsize$\pm.016$} & $0.614${\scriptsize$\pm.013$} & $0.830${\scriptsize$\pm.008$} & {\boldmath$0.843$}{\scriptsize$\pm.003$} & $0.756${\scriptsize$\pm.007$} & $0.755${\scriptsize$\pm.011$} \\
\midrule
\textbf{Mean} & $0.508${\scriptsize$\pm.008$} & $0.457${\scriptsize$\pm.009$} & $0.432${\scriptsize$\pm.003$} & {\boldmath$0.550$}{\scriptsize$\pm.005$} & {\boldmath$0.822$}{\scriptsize$\pm.004$} & $0.644${\scriptsize$\pm.008$} & $0.668${\scriptsize$\pm.005$} & $0.804${\scriptsize$\pm.007$} & \underline{{\boldmath$0.823$}{\scriptsize$\pm.012$}} & $0.736${\scriptsize$\pm.005$} & $0.737${\scriptsize$\pm.009$} \\
\midrule
Fit\,/\,image & --/\tinfSegGPT & --/\tinfUniverSeg & --/\tinfINSID & --/\tinfTyche & \tfitExemplar/\tinfExemplar & \tfitilastik/\tinfilastik & \tfitConvpaint/\tinfConvpaint & \tfitnnUNet/\tinfnnUNet & \tfitCellposeSAM/\tinfCellposeSAM & \tfitStarDist/\tinfStarDist & \tfitmicroSAM/\tinfmicroSAM \\
\bottomrule
\end{tabular}}
\end{table*}

%% file: tab_ablation.tex
\begin{table}[t]
\centering
\caption{Ablation at $K{=}8$, mean $\pm$ standard deviation over 10 seeds of the mean over the datasets. Best per column in bold.}
\label{tab:ablation}
{\scriptsize\setlength{\tabcolsep}{1.2pt}
\begin{tabular}{l ccc}
\toprule
Configuration & Overlap & Centreline & Mean \\
\midrule
Coarse + fine view only & $0.768${\scriptsize$\pm.015$} & $0.847${\scriptsize$\pm.002$} & $0.799${\scriptsize$\pm.009$} \\
Priors only & $0.651${\scriptsize$\pm.016$} & $0.636${\scriptsize$\pm.004$} & $0.645${\scriptsize$\pm.010$} \\
Coarse view + priors & $0.782${\scriptsize$\pm.006$} & $0.737${\scriptsize$\pm.004$} & $0.764${\scriptsize$\pm.004$} \\
Fine view + priors & $0.802${\scriptsize$\pm.007$} & {\boldmath$0.851$}{\scriptsize$\pm.002$} & $0.821${\scriptsize$\pm.004$} \\
\midrule
Last layer only & $0.794${\scriptsize$\pm.007$} & $0.839${\scriptsize$\pm.004$} & $0.812${\scriptsize$\pm.005$} \\
Fine view not magnified & $0.788${\scriptsize$\pm.008$} & $0.763${\scriptsize$\pm.005$} & $0.778${\scriptsize$\pm.005$} \\
\midrule
Plain readout, threshold $0.5$ & $0.798${\scriptsize$\pm.006$} & $0.787${\scriptsize$\pm.004$} & $0.794${\scriptsize$\pm.004$} \\
Balanced readout, threshold $0.5$ & $0.721${\scriptsize$\pm.008$} & $0.815${\scriptsize$\pm.008$} & $0.759${\scriptsize$\pm.006$} \\
\midrule
Forest on the views & $0.703${\scriptsize$\pm.027$} & $0.809${\scriptsize$\pm.005$} & $0.746${\scriptsize$\pm.016$} \\
Forest on the priors & $0.642${\scriptsize$\pm.014$} & $0.578${\scriptsize$\pm.009$} & $0.616${\scriptsize$\pm.008$} \\
\midrule
\textbf{Exemplar} & {\boldmath$0.805$}{\scriptsize$\pm.006$} & $0.848${\scriptsize$\pm.003$} & {\boldmath$0.822$}{\scriptsize$\pm.004$} \\
\bottomrule
\end{tabular}}
\end{table}